\documentclass{article}
\usepackage{ijcai25}

\usepackage{times}
\usepackage{soul}
\usepackage{url}
\usepackage[hidelinks]{hyperref}
\usepackage[utf8]{inputenc}
\usepackage[small]{caption}
\usepackage{graphicx}
\usepackage{amsmath}
\usepackage{amsthm}
\usepackage{booktabs}
\usepackage{algorithm}
\usepackage{algorithmic}
\usepackage[switch]{lineno}

\usepackage{color}
\usepackage{colortbl}
\usepackage{xcolor}

\def \ie {\emph{i.e.}}

\newcommand{\tit}[1]{\smallbreak\noindent\textbf{#1.}}

\newcommand{\lollo}[1]{\textcolor{magenta}{\textbf{#1}}}

\makeatother

\title{The Safety Challenge of World Models for Embodied AI Agents: A Review}

\author{
Lorenzo Baraldi$^{1}$\thanks{Work done during internship at Huawei RAMS Lab.}
\and
Zifan Zeng$^{2,3}$\and
Chongzhe Zhang$^{2,4}$\and
Aradhana Nayak$^2$\and\\
Hongbo Zhu$^{5*}$\and
Feng Liu$^2$\and
Qunli Zhang$^2$\and
Peng Wang$^2$\and
Shiming Liu$^2$\and\\
Zheng Hu$^2$\and
Angelo Cangelosi$^5$\And
Lorenzo Baraldi$^6$\\
\affiliations
$^1$University of Pisa\\
$^2$Huawei RAMS Lab\\
$^3$Technical University of Munich\\
$^4$Technical University of Berlin\\
$^5$University of Manchester\\
$^6$University of Modena and Reggio Emilia\\
\emails
lorenzo.baraldi@phd.unipi.it,
\{zifan.zeng, chongzhe.zhang, aradhana.nayak, hongbo.zhu, feng.liu1, zhangqunli1, wangpeng394, liushiming3, hu.zheng\}@huawei.com,
\{angelo.cangelosi, hongbo.zhu\}@manchester.ac.uk, lorenzo.baraldi@unimore.it
}

\begin{document}

\maketitle

\begin{abstract}
    The rapid progress in embodied artificial intelligence has highlighted the necessity for more advanced and integrated models that can perceive, interpret, and predict environmental dynamics.  In this context, World Models (WMs) have been introduced to provide embodied agents with the abilities to anticipate future environmental states and fill in knowledge gaps, thereby enhancing agents' ability to plan and execute actions. However, when dealing with embodied agents it is fundamental to ensure that predictions are safe for both the agent and the environment.
In this article, we conduct a comprehensive literature review of World Models in the domains of autonomous driving and robotics, with a specific focus on the safety implications of scene and control generation tasks. Our review is complemented by an empirical analysis, wherein we collect and examine predictions from state-of-the-art models, identify and categorize common faults (herein referred to as pathologies), and provide a quantitative evaluation of the results. 

\end{abstract}

\section{Introduction}
\label{sec:introduction}
Recently, autonomous agents such as self-driving cars, humanoid robots and manipulators have been driven by the strong reasoning capability and ease of accessibility to large transformer models. However, researchers frequently try to address the questions, `can training data for an agent be autonomously generated?' and `can an agent reason and perform complex maneuvers in dynamical scenarios?' The common theme underlying these questions is the ability of the agent to comprehend the future evolution of its surroundings from the present state and predict the immediate outcome of its actions. Further, the recent emergence of large language models and their unparalleled levels to linguistic comprehension has led to an increased interest in semantically driven agents.
\begin{figure}
\centering
\includegraphics[width=0.98\linewidth]{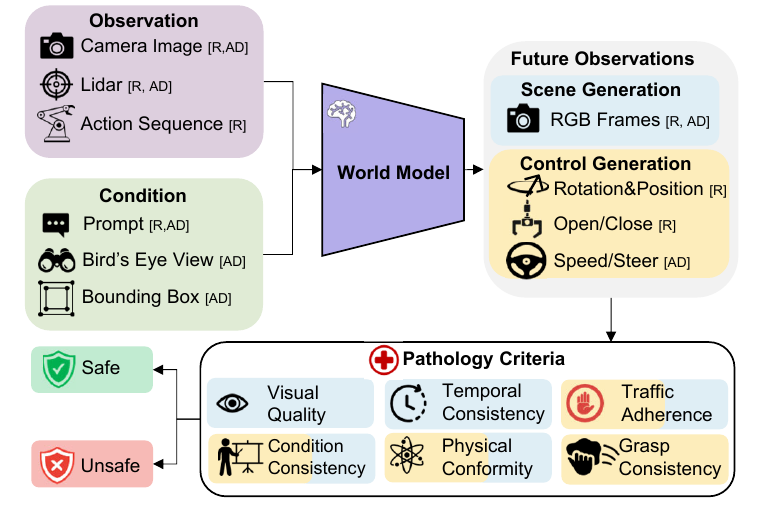}
\vspace{-.2cm}
\caption{Illustration of a World Model in the domain of autonomous driving or robotics. The current observation and conditions are used to predict future observations, on the tasks of novel scenes (yellow) or control actions (blue) generation. Our pathology criteria enable the evaluation of safety in the generated outputs for both tasks.}
\label{fig:world_models}
\vspace{-.4cm}
\end{figure}
This creates a three-fold challenge. The agent must be able to 1) correctly comprehend an instruction 2) process information from its environment in order to interact with participants in it and 3) predict the outcome in the immediate future and thereby take steps to execute the instruction.
To address the first two challenges, researchers have integrated multiple sources of information, derived from perceptual inputs, with the reasoning capabilities characteristic of language models leading to the creation of Multimodal Large Language Models (MLLMs)~\cite{caffagni-etal-2024-revolution}. The third challenge is addressed by so-called World Models (WMs) \cite{cosmos2024} that have been proposed as a potential solution to estimate future environmental observations and possibly missing information. 

Although the concept of WMs is well known in control theory to learn the dynamics of an agent in static or unknown environments, recent research endeavors have significantly advanced WMs development, enabling prediction of future observations and their translation into various forms of output, including videos ~\cite{brooks2024video}, control actions~\cite{robogen}, or a combination of both ~\cite{hu2023gaia}. Through the training process, WMs implicitly acquire a fundamental understanding of the underlying physical laws, dynamics, and spatiotemporal continuity, which is reflected in the generated predictions. However, it is crucial to acknowledge that faulty predictions can lead to catastrophic results in embodied scenarios like autonomous driving. Therefore the focus should not solely be directed to correctly executing the task, but rather on ensuring that the predicted actions and outcomes do not compromise the safety of the agent, its environment, or other entities.

In response to the growing research efforts aimed at constructing better World Models, this literature review undertakes a three-fold objective. Firstly, it provides a comprehensive and systematic overview of the current landscape of World Models in the field of autonomous driving and robotics. Secondly, it conducts an in-depth analysis of the safety aspects of WMs, with a focus on the tasks of scene and control generation. This analysis identifies the faults of world models, which are herein referred to as \textit{pathologies}, as well as it establishes qualitative criteria for their estimation. Finally, it identifies and discusses prominent future research directions for improved safety in WMs.
 
This review differs from existing surveys~\cite{zhu2024sora,guan2024world} in its focus on the safety aspect of WMs, rather than solely on their performance and taxonomy. Further, it builds on the definition proposed in \cite{10771431}, providing detailed analysis on both scene generation and control tasks in the domain of robotics and autonomous driving, including quantitative evaluations on the defined pathology criteria. By doing so, this review aims to provide a comprehensive understanding of the current state of World Models and their safety implications, ultimately contributing to the development of more robust and reliable WMs.

\section{Preliminaries on World Models}
\label{sec:preliminaries}
We define a World Model (Fig~\ref{fig:world_models}) as an entity which accepts the current observation of the environment from sensory modalities along with a condition as inputs and predicts future observations in a short time frame in the form of of one or more sensory modalities. Sensory inputs to the world model are images (camera) or states of a robot (joint torque sensor, cameras). The condition in natural language or a desired future observation in the form of bounding boxes in 2D (bird's eye view with a drone) is a stimulus to change the current observation that is provided to the WM by a user or by a task organizer which dispatches goals valid for the small time frame. In what follows we discuss notable works in two broad domains of utilization of WMs: scene generation and control. Within each category, we explore two application areas: autonomous driving and robotics.

\subsection{World Models for Scene Generation}
\begin{table}
\centering
\setlength{\tabcolsep}{.15em}
\resizebox{\linewidth}{!}{
\begin{tabular}{l c c c cl}
\toprule
\textbf{Model} & \textbf{Base Model} & \textbf{Output-Resolution} & \textbf{Task} \\
\midrule1

GAIA-1 \cite{hu2023gaia} & D + A & $(1\times288\times512)\times26$ & AD \\

DriveDreamer \cite{wang2023drive} & D + A & $(1\times256\times448)\times32$ & AD \\

DriveDreamer2 \cite{zhao2024drive} & D + LLM & $(6\times256\times448)\times8$ & AD \\

MagicDrive \cite{gao2023magicdrive} & D & $(6\times224\times400)\times7$ & AD \\

Panacea \cite{wen2024panacea} & D & $(6\times256\times512)\times8$  &  AD \\

Drive-WM \cite{wang2024driving} & D & $(6\times192\times384)\times8$ &  AD \\

DrivingDiffusion \cite{li2025drivingdiffusion} & D & $(6\times512\times512)\times6$  & AD \\

DreamForge \cite{mei2024dreamforge} & D & $(6\times224\times400)\times7$ &  AD \\

DriveScape \cite{wu2024drivescape} & D & $(6\times576\times1024)\times30$ & AD \\

Vista \cite{gao2024vista} & D & $(1\times576\times1024)\times25$ &  AD \\

MagicDriveDiT \cite{gao2024magicdrivedit} & D & $(6\times848\times1600)\times129$ & AD \\

\midrule
AVDC~\cite{ko2023learning} & D & $(1\times48\times64)\times7$ &  R \\
EVA~\cite{chi2024eva} & D + LLM & $(1\times256\times256)\times16$ &  R \\
VideoAgent~\cite{soni2024videoagent} & D & $(1\times48\times64)\times7$ &  R \\
This\& That~\cite{wang2024language} & D & $(1\times256\times384)\times25$ &  R \\

\midrule

Open-Sora Plan~\cite{lin2024open} & D & $(1\times352\times640)\times93$ & G \\
Cosmos~\cite{cosmos2024} & D + A & $(1\times704\times1280)\times120$ &  G \\

\bottomrule
\end{tabular}
}
\vspace{-.15cm}
\label{tab:scene_generators}
\caption{
A summary of world models for scene generation. Base Model indicate, D: Diffusion, A: Autoregressive. Possible tasks are AD: Autonomous Driving, R: Robotics, G: General Purpose
\vspace{-.38cm}
}
\end{table}
The scene generation task involves predicting future observations in the form of sequential video frames.
WMs for scene generation typically rely on two primary generative frameworks: autoregressive or diffusion models. Autoregressive models~\cite{hu2023gaia,cosmos2024} generate future observations in a sequential manner, conditioning each new prediction on the previous ones. In contrast, diffusion models~\cite{gao2024magicdrivedit,zhu2024sora} operate by iteratively refining a Gaussian noise signal, progressively removing noise to produce a coherent visual representation.

In what follows, we will focus mostly on diffusion models~\cite{dhariwal2021diffusion}, which have emerged as a prominent architectural framework in both the domains of autonomous driving and robotics scene generation.

\tit{Autonomous Driving}
GAIA-1~\cite{hu2023gaia}, is one of the pioneering World Model for driving scene generation and it operates on two components: a WM and a video diffusion decoder. The World Model is in charge of high-level reasoning about the subjects represented in the scenes and their dynamics. Specifically, it tokenizes multimodal inputs (\ie video, text, action) into discrete tokens and autoregressively predicts the next observation in the latent space. The diffusion decoder, instead, is in charge of rendering the video based on the generated tokens.  
Differently, in DriveDreamer~\cite{wang2023drive} the autoregressive module is only used to predict future actions, while the conditioning input is expanded to HDMap, 3D Bounding box, initial frame and textual prompt, to improve the consistency on generated video. 
Further, DriveDreamer2~\cite{zhao2024drive} removes the autoregressive module and eliminates the need for 3D Boxes and HD Maps from real images by generating them via LLM using user prompts. 
Unlike methods that explicitly provide position information for objects at every moment, Vista~\cite{gao2024vista} is conditioned on three initial frames, allowing the diffusion module to learn to capture the motion priors of objects and generate plausible movements.

On the other hand, MagicDrive \cite{gao2023magicdrive} and Panacea \cite{wen2024panacea} implement video generation for driving scenes with six views (front-left, front-center, front-right, back-left, back-center, and back-right) 
The generation is conditioned on both text prompts and sequences of Bird Eye Views (BEVs), and is conducted in two stages: firstly, generating multi-view images, and secondly, using these images to generate future frames.
To improve the consistency and realism of objects across views in the generated video, DrivingDiffusion~\cite{li2025drivingdiffusion} and DreamForge~\cite{mei2024dreamforge} enhance the local attention to key objects.

Differently, DriveScape~\cite{wu2024drivescape} adopts a novel approach by transforming the process of generating six views at once into a two-step procedure: first, freely generating three non-overlapping perspectives, and then considering the spatial coherence between adjacent views to generate the remaining three views.
In terms of boosting the length of generated videos, MagicDriveDit~\cite{gao2024magicdrivedit} makes a significant contribution by converting the architecture from the original U-Net implementation to the more scalable Diffusion Transformer~\cite{peebles2023scalable} architecture, which overcomes single GPU memory constraints due to sequence parallelism.
Beyond generating driving videos, Drive-WM~\cite{wang2024driving} applies the generated videos to the training of an autonomous driving planner and improves the planner's capabilities in out-of-distribution scenarios.

\tit{Robotics}
In AVDC~\cite{ko2023learning}, the authors introduce a WM that generates manipulator actions. Firstly a diffusion model~\cite{dhariwal2021diffusion} synthetizes a video representing the required actions that are subsequently translated in movements of the  manipulator via optical flow estimation. Following this method, VideoAgent~\cite{soni2024videoagent}, improves AVDC by including feedbacks of a Multimodal Large Language Model (MLLM) to refine the video generation pipeline, making it more consistent. 
In contrast, This\&That~\cite{wang2024language} focuses on addressing the ambiguity of textual prompts in robotics diffusion model generators. This is achieved by simplifying the textual prompt to pointing gestures and words, such as \texttt{put this pen on that table}. The gestures are then translated to 2D coordinates in the initial frame and employed as additional conditioning for the generation process.

An alternative approach is presented in EVA ~\cite{chi2024eva}, where an MLLM is iteratively prompted to generate a video that solves a given task, based on current observation and task description. The MLLM refines its initial video prediction through multiple iterations until task completion is achieved (which is estimated by the model itself). 

\tit{General Purpose} The research domain of video generation extends far beyond the specific application of robotics and driving scene generation.
Cosmos~\cite{cosmos2024} comprises a family of world foundation models (including both diffusion-based and autoregressive architectures) which have been trained on a large-scale video dataset consisting of approximately 100 million clips. These models are capable of generating videos conditioned on either text-only inputs or a combination of text and previous visual observations. Further, Cosmos optimizes the latent space representation of videos, enabling more efficient and effective video generation.
In a similar vein, Open-Sora Plan~\cite{lin2024open} is a diffusion model generator that targets the video encoding pipeline and the diffusion transformer architecture.

\subsection{World Models for Control}
\begin{table}
\centering
\setlength{\tabcolsep}{.05em}
\resizebox{\linewidth}{!}{
\begin{tabular}{l c c c cl}
\toprule
\textbf{Model} & \textbf{Base Model} & \textbf{Input Modality} & \textbf{Task} \\
\midrule

MILE~\cite{hu2022mile} & RNN & Im, M & AD \\

CTG++~\cite{zhong2023ctgplus} & D & Tr, M, Tx & AD\\

LCTGen~\cite{tan2023lctgen} & MLP & Tr, M, Tx & AD\\

GenAD~\cite{zheng2024genad} & RNN & Im, M & AD\\

SEM2~\cite{gao2024sem2} & RNN & Im, Pc & AD \\

Think2Drive~\cite{li2024think2drive} & RNN & Tr, M & AD \\


Doe-1~\cite{zheng2024doe} & A & Im, Tr, Tx & AD \\

GUMP~\cite{hu2024gump} & A & Tr, M & AD \\

\midrule
BC-Z \cite{bcz} & CNN & Im, Tx &  R \\
RT-1 \cite{rt1} & A & Im, Tx &  R \\
SayCan \cite{saycan} & CNN+LLM & Im, Tx &  R \\
PaLM-E \cite{palme} & MLLM & Im, Tx &  R \\
Eureka \cite{eureka} & MLP+LLM & S, Tx &  R \\
Octo \cite{octo} & CNN+A & Im, Tx &  R \\
Robogen \cite{robogen} & MLP+LLM & S, Tx &  R \\
OpenVLA \cite{openvla} & LLM & Im, Tx &  R \\
Manipulate Anything \cite{manipulateanything} & MLLM & Im, Tx &  R \\
\bottomrule
\end{tabular}
}
\vspace{-.15cm}
\caption{\label{tab:control}
A summary of world models for control generation. D: Diffusion, A: Autoregressive; Im: Image, Tr: Trajectory, M: Map, Pc: Pointcloud, Tx: Text, S: Simulator Environment. 
\vspace{-.38cm}
}
\end{table}
WMs presented in this subsection predict future observations in the form of sparse control signals.
In autonomous driving, WMs can be used to control the ego vehicle in closed-loop or to control multiple background vehicles generating traffic scenarios. 
In robotics, WMs are used to control the agent with an initial image (encapsulating the current environmental state) and the desired goal (in the form of image/language instruction). In what follows, we describe SoTA methods for control using WM for autonomous driving and robotics.

\tit{Autonomous Driving} WMs for control in autonomous driving are primarily divided into two applications: controlling the ego vehicle for closed-loop driving and generating the future trajectories of background vehicles based on the current state to create traffic scenarios. For the former, MILE~\cite{hu2022mile} is a model-based Imitation Learning (IL) approach with a Recurrent Neural Network (RNN)~\cite{rumelhart1988rnn} in latent space and an encoder-decoder architecture to process vision input. MILE is the first method that realizes closed-loop autonomous driving with only visual input in urban scenarios of the Carla simulator~\cite{dosovitskiy17carla}. 
Think2Drive~\cite{li2024think2drive} adopts a Recurrent State-Space Model (RSSM)~\cite{hafner2019learning} instead of the RNN, to predict the transition of environment states in latent space for the planner to give control signals. The improved method is the first model that successfully completes all 39 challenging scenarios in Carla Leaderboard v2. 
On the other hand, GenAD~\cite{zheng2024genad} and SEM2~\cite{gao2024sem2} utilize Gated Recurrent Unit (GRU)~\cite{cho2014gru} to predict the future states in the latent space. The difference is that GenAD uses self- and cross-attention mechanisms to process temporal and multimodal information, whereas SEM2 proposes to train a semantic latent filter to mask out the irrelevant information in the latent space. With the introduction of the transformer, Doe-1~\cite{zheng2024doe} considers closed-loop autonomous driving as a next-token generation problem and trains a multi-modal autoregressive transformer to generate tokens for the tasks of visual question answering, action-conditioned video generation, and end-to-end motion planning as a unified framework. 

In traffic scenario generation, CTG++~\cite{zhong2023ctgplus} trains a diffusion model to sample the trajectory of vehicles from Gaussian noise. To improve the controllability of the generation, a differentiable target function is introduced during the denoising process to guide the generation explicitly through SGD. With the help of LLM, the target function can be designed using natural language to describe the guidance of the generated traffic scenario. 
Differently, in LCTGen~\cite{tan2023lctgen} and GUMP~\cite{hu2024gump}, transformer architecture is introduced to predict the embeddings of the states of the environment for the next few steps. Similar to CTG++, LCTGen incorporates an LLM-driven text interpreter to process language input as the condition for the traffic scenario.

\tit{Robotics}
WMs can be used to generate trajectories to control an embodiment such as a manipulator, a mobile base or a humanoid robot. WM predicts the pose of the end-effector of a manipulator (or generates \textit{actions}) in 6-DOF space, in 1-DOF for a gripper, in 2-DOF for a mobile base,  or in the joint space of a humanoid agent. In robotics, WMs have been used in context of language conditioned IL methods (also called transfer learning) such as BC-Z \cite{bcz} and RL methods such as MT-OPT \cite{mtopt}, wherein the focus is learning the dynamics of the embodiment along with the tasks. These methods encourage task, environment and embodiment generalizability and the research challenge is to learn from few \textit{demonstrations} in IL and from few trials in RL. \textit{Demonstrations} are robot trajectories labelled with task instructions generated by an expert through tele-operation or by kinesthetic teaching and hence, they are expensive to obtain. Furthermore in these methods, the network architecture is designed to either maximize a reward function in case of RL or to determine a policy that matches demonstration data as closely as possible in case of IL. With the ease of availability in MLLMs, there is an increased effort in recent years to train IL policies on this architecture such as Octo \cite{octo}, RT-1 \cite{rt1}, OpenVLA\cite{openvla} along with generation of massive amount of demonstration data such as Open X-Embodiment dataset \cite{openx}. In RT-1, vision language \lollo{.} tokens are generated via CNN and a transformer outputs \textit{action} tokens. Octo Base (93M parameters) is trained on a large transformer, it accepts goal images instead of task instructions as inputs and performs better than RT-1 trained on Open X-Embodiment (RT-1-X) in zero shot evaluation. OpenVLA predicts end effector and gripper \textit{actions} by projecting visual features into language embedding space followed by an LLM backbone and it shows better generalizability compared to Octo and RT-1. 

Since the generalizability of RL approaches to new tasks is better compared to IL \cite{fu2018from} and, given excellent code generation capability of LLMs, many approaches focus on reward code generation from language instruction RoboGen~\cite{robogen}, Eureka~\cite{eureka}, and Manipulate Anything~\cite{manipulateanything}. In RoboGen \textit{demonstrations} are generated in a simulator environment by first defining sub-tasks followed by policy generation through RL or sampling based motion planning. In Eureka~\cite{eureka} reward function code is generated from simulator environment source code through an LLM and prompts. In Manipulate Anything~\cite{manipulateanything}, sub-tasks are verified allowing for possible error recovery through language conditions and, a MLLM generates \textit{action} code through prompts and task primitives. Given a long horizon task, some approaches use LLM for task and motion planning from a vocabulary of skills or language labelled \textit{demonstrations}. PaLM-E \cite{palme} generates sub-tasks auto-regressively through multi-modal inputs to an LLM. SayCan \cite{saycan} chooses the right skill sequence with an LLM from a value function space of skills with language descriptions called `affordance space'. Safety constraints from natural language can be possibly incorporated in this space. 

\section{Pathologies in World Models Safety}
\label{sec:pathologies}
Our study endeavors to address the safety limitations of World Models. To this end, we undertake a two-step approach. Firstly, we generate future observations with state-of-the-art WMs focusing on the scene generation and control tasks. Secondly, we perform a thorough analysis of the generated data, with a primary focus on identifying and categorizing anomalous scenarios, that we refer to as \textit{pathologies}. To facilitate this analysis, we establish a set of \textit{pathology criteria} to classify the faults of WMs, providing a systematic framework for future research.

\subsection{Experimental Settings}
Concerning scene generation we include in our analysis WMs that were expressly designed for the generation of driving scenes, namely Panacea~\cite{wen2024panacea}, Vista~\cite{gao2024vista}, and MagicDriveDit~\cite{gao2023magicdrive}. Additionally, we employ Cosmos~\cite{cosmos2024} and Open-Sora Plan~\cite{lin2024open} as general purpose scene generators and This\&That~\cite{wang2024language} for robotics scenes.  General purpose and AD WMs are tasked to generate driving scenes from the nuScenes~\cite{nuscenes} validation set. In cases where models require textual prompts as condition, we prompted ChatGpt-4o-mini to generate descriptive caption based on the initial frame.  Differently we prompt This\&That to generate 100 scenes of the Bridge-v2 dataset \cite{walke2023bridgedata}. We follow This\&That input pipeline to transform the textual task in visual coordinates.
\begin{figure*}
\centering
\includegraphics[width=0.988\linewidth]{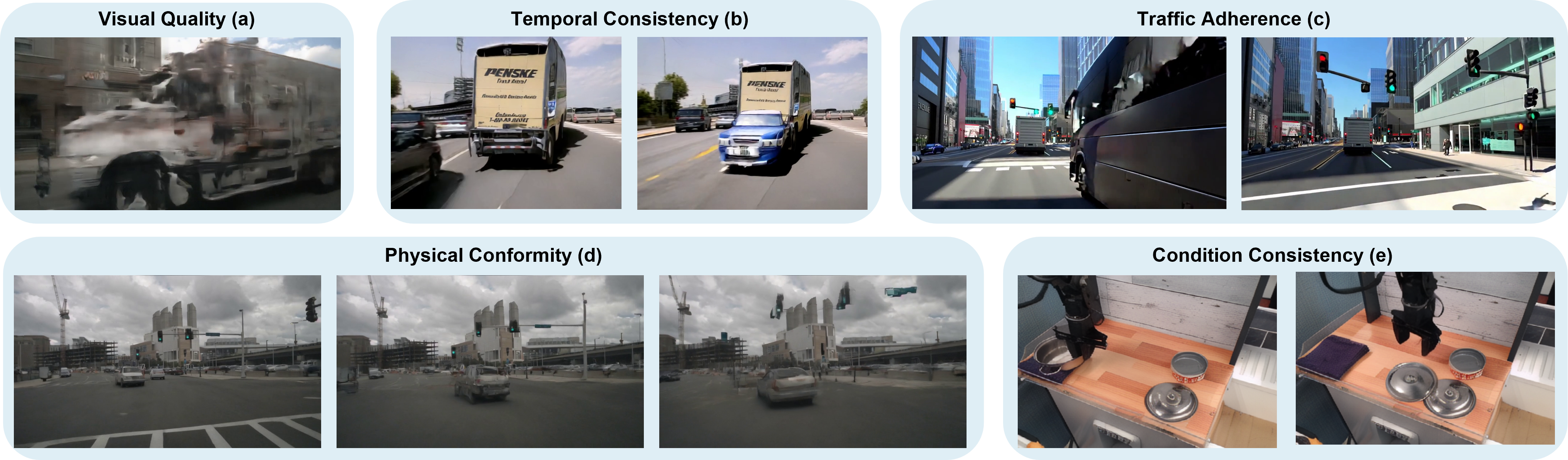}
\vspace{-.2cm}
\caption{Illustration of pathologies identified in SoTA WMs for scene generation. Visual Quality (MagicDriveDit), Temporal Consistency (Open-Sora), Traffic Adherence (Comsos), Physical Conformity (Vista), Condition Consistency (This\&That)}
\label{fig:criteria_generation}
\vspace{-.4cm}
\end{figure*}

Regarding WMs for AD control, we select MILE~\cite{hu2022mile}, CTG++~\cite{zhong2023ctgplus}, and LCTGen~\cite{tan2023lctgen}. As traffic scenario generator, CTG++ and LCTGen take the initial frames of 100 randomly selected scenarios from the nuScenes~\cite{nuscenes} dataset and the Waymo Open Dataset~\cite{sun2020waymoopendataset}, respectively. To verify the safety of interactions between multiple vehicles controlled by MILE, we deploy the models on multiple agents within the same simulation scenario in Carla. We adopt the method proposed in \cite{li2024rigorous} to setup challenging scenarios and explore edge configurations of the initial states.

In robotics control, we consider 3 SoTA models RT-1-X \cite{rt1} and Octo-Base \cite{octo} trained on Open-X dataset and, RoboGen \cite{robogen}. Our choice is motivated by their different base architectures, parameters and training. The tabletop set-up for RT-1-X, Octo is that of Bridge-v2 dataset (WidowX arm with parallel gripper) realised in a OpenAI Gym Simulator through SimplerEnv \cite{simpler}. For RoboGen, a Franka 7-DOF arm mounted on a mobile base, 2 finger Franka Hand parallel gripper are used and tasks are set on tabletop in PyBullet simulator. All the example tasks in Bridge-v2 provided in SimplerEnv are tested along with 10 new instruction set-up by introducing 1) positional changes in language instruction (e.g. `outside' to `inside') and 2) sub-tasks by altering the orientation of the manipulated object. We test RoboGen in 4 example tasks and 7 new tasks involving 1) sensitive objects (eg. laptop, phone) 2) positional language instruction such as `\texttt{Place bottle \textbf{beside} laptop}'.

\subsection{Pathology Criteria in WM for Scene Generation}
\tit{Visual Quality} Building on the definition of VideoScore~\cite{he2024videoscore}, a relevant pathology criteria for generated scenes consist on assessing the representations of all the depicted subjects, verifying their clarity, shape, and color correctness. 
A violation of this criterion would result in blurred frame regions, unrealistic shapes and deformed objects e.g. in Fig~\ref{fig:criteria_generation} (a), wherein a truck is in an odd shape.

\tit{Temporal Consistency}
Temporal consistency~\cite{liu2024evalcrafter} assess the preservation of subject continuity across generated frames. In details, each represented subject should maintain a consistent visual identity (shape, color, and texture) throughout the generated video, while the scene should exhibit an adequate level of dynamism (avoiding static representation). 
This criterion should be enforced also in case of multi-view scenes generation (e.g frontal and back cameras in autonomous driving) where subjects that leave the field of view of a camera could enter in another one in case of coherent motion.
Any deviation, such as  abrupt introduction, removal, or change of visual entities, constitute a violation of temporal consistency for e.g. in Figure~\ref{fig:criteria_generation} (b) the blue car abruptly appears from the back of the truck.

\tit{Traffic Adherence (AD only)}
To ensure the generation of realistic and safe driving scenarios, it is essential that the produced scenes conform to standard traffic regulations, unless explicitly prompted to deviate from them. For instance, at intersections regulated by traffic lights, vehicles should remain stationary when the signal is red, and proceed when the signal turns green. Further, any vehicle maneuver, such as turning, merging, or changing lanes, should be informed by the presence of other subjects in the nearby area. 
A violation of the criteria is illustrated in Fig~\ref{fig:criteria_generation} (c), where a truck is depicted crossing a road with an inconsistent traffic signal sequence, specifically a red light followed by a green light. 

\tit{Physical Conformity}
Generated scenes must adhere to the fundamental principles of physics to ensure safety.  For instance, the motion of represented subjects should exhibit smooth and continuous trajectories, avoiding abrupt and anomalous movements~\cite{Huang_2024_CVPR}. 
Further, the generated scenes should also adhere to the principles of gravity, avoiding floating objects and, in the event of a direct collision between two objects, the generated scene should accurately depict the resulting interactions. Violations of physical conformity criteria is illustrated in Fig~\ref{fig:criteria_generation} (d), where a car undergoes an abrupt change in its driving trajectory, modifying its orientation between the first and second frames. The third frame violates gravity: traffic lights are floating in the air.

\tit{Condition Consistency}
Condition consistency is concerned with assessing the degree to which generated scenes conform to the conditioning information. While current literature focus mostly on text-to-video alignment~\cite{he2024videoscore}, conditioning information can take various forms, including but not limited to: textual prompts, LiDAR data, images, videos, 2D/3D bounding boxes, trajectories, bird's eye views. 
A violation of the criterion is illustrated in Fig~\ref{fig:criteria_generation} (e), where the user's prompt `\texttt{Take the red object out of the pot and put it on the left burner}' is not fulfilled. Instead, only the lid is placed on the burner without extracting the red object which is not present in the pot.

\begin{figure}
\centering
\includegraphics[width=0.9\linewidth]{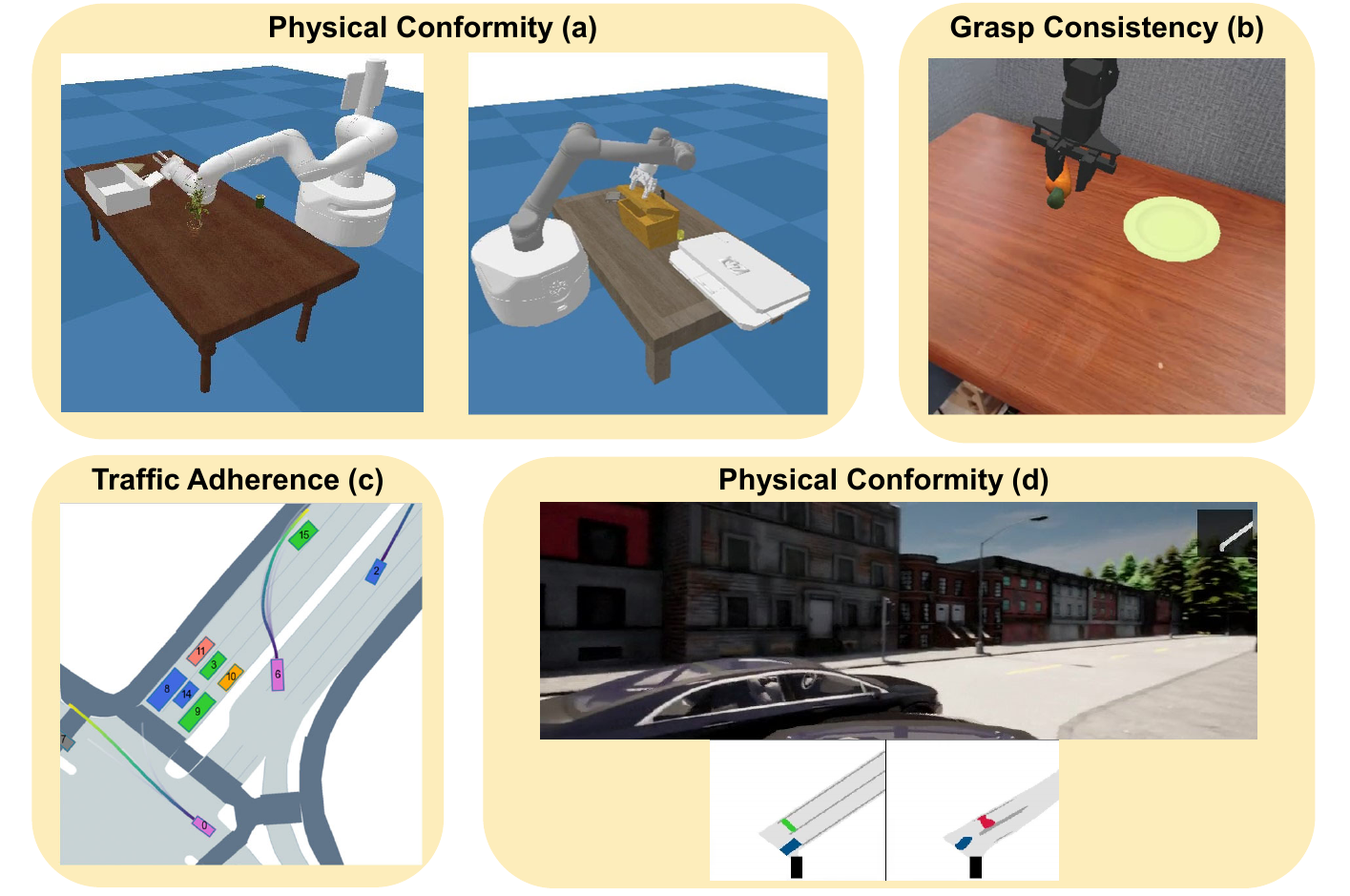}
\vspace{-.2cm}
\caption{Illustration of pathologies identified in SoTA WMs for control: RoboGen (a), Octo (b), MILE (c), (d)}
\label{fig:criteria_generation_control}
\vspace{-.4cm}
\end{figure}

\subsection{Pathology Criteria in WM for Control}
\tit{Condition Consistency} For robotics, we refer to two aspects of input conditioning which is applied through the user provided instruction (or prompt). The first refers to recognizing the correct articulated object amidst distractors. The second refers to task comprehension and thereby task completion through the generated control sequence. In experiments we observed that models had poor positional awareness hence failed in tasks and, also collided with nearby objects.

Concerning ego vehicle controller, condition consistency refers to completing the navigation task and traveling to the goal in the scenarios. Differently, for traffic scenario generation, condition consistency assess whether models can generate trajectories of the background vehicles that are consistent with the language condition.

\tit{Physical Conformity}
The generated policy of WM should avoid a direct impact on the embodiment in both AD and R. In R, we observe that the tested models avoid self-collision. In those trained on demonstration data, sensitive manipulation is not an implicit feature. Motion planning based on sampling approaches which are suited for navigation tasks suffer from possible collision occurrence within a time-step when used for manipulation e.g. collision of gripper and base with table in Figure \ref{fig:criteria_generation_control} (a) and (b) respectively. In AD, the safety-relevant criterion described by physical conformity is the collision with surrounding vehicles e.g. in Figure \ref{fig:criteria_generation_control} (d), where the ego vehicle controlled by MILE collides into another vehicle which is controlled by another MILE instance.

\tit{Grasp Consistency (R only)}
The generated grasp poses must respect physical characteristics of the gripper and correspond to the manipulated object. The gripper should close after the correct grasp pose and arm position are both attained. Lack of coordination leads to task failure e.g. in Fig~\ref{fig:criteria_generation_control} (b). 

\tit{Traffic Adherence (AD only)} The selection criterion for the generated control trajectories that disobey traffic rules mainly focuses on the violation of driving direction and the drivable area for e.g. in Fig~\ref{fig:criteria_generation_control} (c), where vehicle 6 crosses the boundary between the lanes in different directions and travels into the reverse lane.

\section{Metrics for World Models Safety}
\label{sec:metrics_benchmarks}
\tit{Metrics for Scene Generation}
A popular evaluation metric for World Models for scene generation consist of the Fréchet video distance (FVD) \cite{unterthiner2019fvd}. However, FVD has been proven to favor frame quality against temporal realism~\cite{ge2024content}. While this problematic has been addressed by employing a different feature extractor backbone, it is clear that FVD can not provide detailed insights on each of the pathology enumerated in Section~\ref{sec:pathologies}.

The adoption of MLLMs has recently pushed the performance of AI-generated content evaluation. Notably, \cite{he2024videoscore} fine-tunes a MLLM to evaluate generated videos along multiple dimensions, including visual quality, temporal consistency, dynamic degree, text-to-video alignment, and factual consistency. Similarly, \cite{qin2024worldsimbench} employs an MLLM to evaluate autonomous driving, robotics, and open-ended embodied scenes, taking into account different criteria and collecting 35k human annotations for training purposes. Differently, in~\cite{arai2024act} the authors propose a pipeline to evaluate condition consistency in autonomous driving generated scene, by building an estimator module  that compares the prompted driving commands with the actual generated trajectory. 

\tit{Metrics for Control in AD}
Concerning the generation of traffic scenarios, evaluation metrics for trajectory prediction tasks are commonly used, such as Maximum Mean Discrepancy (MMD), mean Average Distance Error (mADE), mean Final Distance Error (mFDE), which are not related to safety, though. CTG++~\cite{zhong2023ctgplus} proposes to use assessment functions derived from Signal Temporal Logic (STL) formulas to calculate the violation of traffic rules, which can be adopted to measure the level of traffic adherence as introduced in Section \ref{sec:pathologies}. 
Differently, the Scenario Collision Rate (SCR) and Success Rate (SR) are mostly employed to evaluate the performance in closed-loop autonomous driving for WMs that control the ego vehicle. The former evaluates whether the vehicle driven by the WM follows physical conformity and avoids collision, and the latter indicates if the model can complete the conditioned driving task. 

\tit{Metrics and Benchmarks for Robotics}
Most WM robotics models focus on generalization to unseen tasks, environment and embodiment through few-shot or zero-shot learning methods. There exist metrics for high-level task planning by WM agent such as LoTa-Bench, Behavior 1K, or for policy generation with RL such as Meta-world \cite{pmlr-v100-yu20a} where focus is on condition consistency and generalization capability. SimplerEnv \cite{simpler} is a simulator-based assessment for both high level plans and \textit{actions}. The evaluation process identifies if the correct or incorrect object was moved or grasped, grasp success (grasp consistency) and finally if correct object reaches the target (both aspects of condition consistency) for a limited set of instructions. The evaluations are based on contact information, object poses and scene information provided by the simulator. ManipulateAnything \cite{manipulateanything} breaks down the type of errors to perception or reasoning (high level task execution capability) and quantifies them by a human-in-loop. There is limited attention in the literature in developing metrics to evaluate all the \textit{pathology criteria} identified in Section~\ref{sec:pathologies} for arbitrary natural language tasks. 

\section{Quantitative Results}
\begin{table}
\centering
\setlength{\tabcolsep}{.35em}
\resizebox{\linewidth}{!}{
\begin{tabular}{l c c c c c c c c l}
\toprule
& \multicolumn{3}{c}{\textbf{VideoScore}} & & \multicolumn{2}{c}{\textbf{Qwen-72B}} \\
\cmidrule{2-4}
\cmidrule{6-7}
\textbf{Model} & \textbf{V} & \textbf{TC} & \textbf{C} & & \textbf{P} & \textbf{TA} & \textbf{AVG} \\
\midrule

Panacea \cite{wen2024panacea} & 2.3 & 2.0 & 3.1 & & \textbf{3.0} & \textbf{2.7} & 2.6  \\

Vista \cite{gao2024vista} & 2.4 & 1.9 & 3.1 & & 2.3  & 1.9 & 2.3  \\

MagicDriveDiT \cite{gao2024magicdrivedit} & \textbf{2.6} & \textbf{2.2} & \textbf{3.2} & & 2.9  & 2.4 & \textbf{2.7}  \\

This\&That\cite{wang2024language}& 2.2 & 1.8 & 3.0 & & \textbf{3.0}  & - & 2.5  \\

\midrule

Open-Sora Plan~\cite{lin2024open} & 2.3 & 2.0 & 3.1 & & 2.5  & 2.2 & 2.4 \\

Cosmos~\cite{cosmos2024} & 2.3 & 1.9 & 3.1 & & 2.9  & 2.5 & 2.5 \\

\bottomrule
\end{tabular}
}
\vspace{-.15cm}
\caption{\label{tab:quantitative}
Quantitative evaluation of scene generation. Visual Quality (V), Temporal Consistency (TC), and Condition Consistency (C) are estimated with VideoScore. Physical Conformity (P) and Traffic Adherence (TA) are estimated with Qwen-72B model. 
\vspace{-.20cm}
}
\end{table}

\begin{table}
\centering
\setlength{\tabcolsep}{0.65em}
\resizebox{\linewidth}{!}{
\begin{tabular}{l c c c c c c c cl}
\toprule
& \multicolumn{3}{c}{\textbf{Examples}} & & \multicolumn{3}{c}{\textbf{New Instructions}} \\
\cmidrule{2-4}
\cmidrule{6-8}
\textbf{Model} & \textbf{C-I}  & \textbf{C-II} & \textbf{G} && \textbf{C-I}  & \textbf{C-II} & \textbf{G} \\
\midrule

Octo~\cite{octo} & 47.9 & 16.0 & 46.5 & & 16  & 0 & 9.3 \\

RT-1-X~\cite{rt1} & 16.5 & 1.7 & 9.1 && 29.3 & 4 & 14.7 \\

RoboGen~\cite{robogen} & 100 & 50 & 50 && 100 & 71 & 14 \\

\bottomrule
\end{tabular}
}
\vspace{-.15cm}
\caption{\label{tab:quantitative_control_robotics}
Quantitative evaluation of robot control. Condition consistency C-I (moved correct object), C-II (sourced object on target) and grasp consistency: G (sourced object grasped) percentages.
\vspace{-.38cm}
}
\end{table}
\tit{WMs for Scene Generation} We adopt VideoScore~\cite{he2024videoscore} to assess with a score 1-4 the pathologies of visual quality, temporal consistency, and condition consistency (text-to-video alignment in our experiments). Further, we score the remaining criteria (~\ie physical consistency and traffic adherence) with the Qwen-72B MLLM \cite{wang2024qwen2} following a similar prompt of VideoScore. However, while MLLM metrics can serve as a proxy for performance, human annotation remains the most reliable method for addressing pathologies of generated scenes. The evaluation results averaged over all generated scenes are reported in Tab~\ref{tab:quantitative}. 

Looking at the table, we have to consider the mutual influence of multiple factors: the length of the generated video, the input condition and the output resolution. 
Theoretically, shorter video sequences are expected to yield better results, as longer generation sequences are more susceptible to error propagation. This hypothesis is supported by the average score of 2.6 achieved by Panacea, which generates relatively short sequences of 8 frames. Conversely, models that rely solely on image frames and textual descriptions, as conditioning, may be more susceptible to performance degradation when tasked with generating longer sequences of frames compared to having additional conditioning constraint like bounding box and BEV representation. This phenomenon is particularly relevant for Vista, Open-Sora Plan, and Cosmos with averages a score of 2.3, 2.4 and 2.5 respectively. On the other hand, additional conditioning information and architectural change in MagicDriveDit favor this model in on this experimental setting. Notably, the average score of all models falls below 3, highlighting the existing gap between current performance and realistic generations (score 4).

\tit{WMs for Control} We evaluate condition and grasp consistency through functions provided in SimplerEnv \cite{simpler}. We test Octo in 288 and RT-1-X in 119 example sub-tasks of the type `\texttt{Place object1 on object2}' based on two aspects of condition consistency: C-I (moved correct object), C-II (sourced object on target) and grasp consistency: G (sourced object grasped).  We also test Octo and RT-1-X in 75 new instruction sub-tasks of the type, `\texttt{Place object1 <preposition> object2}'. We observe that even if Octo performs better in pathologies considered for example tasks, in new instructions RT-1-X exceeds Octo in all parameters. For `\texttt{Cube right of Plate}', RT-1-X successfully grasps and executes 3 of 15 sub-tasks while Octo fails. In 11 manually evaluated tasks of RoboGen, the grasp inconsistency is due to change from parallel to suction in gripper functionality, physical consistency is violated in $45.5\%$ of tasks (and, several times per task) as opposed to isolated occurrences in $5\%$ for both Octo and RT-1-X. Misplaced objects (eg. lamp on floor, laptop on the table etc.) are also generated along with unnecessary sub-tasks (eg. open and close box for a task description `\texttt{Place phone near the box}').

\section{Conclusions and Future Directions}
\label{sec:future_research}
In this article we present a literature review of World Models, with an emphasis on the safety implications of their predictions. Further, we have identified and elucidated key criteria for defining pathologies in the generation, through an analysis of future observations generated by SoTA World Models. Building on this foundation, we proceed to outline promising research directions that aim to address detected pathologies.

\tit{Improved Metrics} Concerning scene generation, physical conformity has only been addressed by explicitly prompting generators to depict physics-related event~\cite{meng2024towards} which is not applicable to the tasks of AD and robotics where physics is an intrinsic aspect of the video. Similarly, traffic adherence has been investigated in \cite{qin2024worldsimbench} where generated videos are translated to control signals, and driving actions are evaluated inside CARLA simulator.

\tit{Self-Improving Methods} Recent studies have demonstrated the potential of MLLMs used as feedback for the scene generation process \cite{soni2024videoagent}, either for fine-tuning or as a refiner at inference. Further, iterative verification based on semantic metrics can improve grasp and condition consistency \cite{aicmllm,manipulateanything} in robotics.

\tit{Neurosymbolic AI for Control Generation} WMs for control generation could leverage rigorous symbolic logic~\cite{nsai} to build the guardrail of the control signals for embodied agents. The limitation in generazibilty of existing Neurosymbolic methods in unseen scenarios could be mitigated by the development of MLLM techniques, serving as a potential bridge between unstructured data and the semi-formal expression of safety-related symbolic knowledge.

\tit{Correction in Affordance Space} The identified pathologies in robotics arise from poor understanding of the affordance space which has been addressed for grasping consistency in \cite{moka}, for physical consistency in \cite{voxposer} by breaking down the end-to-end \textit{action} generation to first generating high level plans followed by low level reasoning in affordance space through value function or keypoints. 

\tit{Reward Function Design} In methods such as RoboGen where policy is derived through RL, the backbone predicts reward functions. 
However, while the high level plan for user prompt does take into account sensitivity of articulated objects, the reward function of the sub tasks is defined simply as vector norm of difference in poses. Sophisticated reward function from language such as \cite{eureka} offer a promising research direction.

\bibliographystyle{named}
\bibliography{ijcai25}

\end{document}